\documentclass[10pt,twocolumn,letterpaper]{article}

\usepackage{wacv}
\usepackage{times}
\usepackage{epsfig}
\usepackage{graphicx}
\usepackage{amsmath}
\usepackage{amssymb}
\usepackage[font=footnotesize]{caption}
\usepackage{subcaption}
\usepackage{url}
\usepackage{ragged2e}

\wacvfinalcopy 
\def\wacvPaperID{} 

\ifwacvfinal\fi
\begin{document}


\title{Adaptive Target Recognition: A Case Study Involving Airport Baggage Screening }

\author{     Ankit Manerikar
\hspace{2cm} Tanmay Prakash 
\hspace{2cm} Avinash C. Kak \\
School of Electrical and Computer Engineering, Purdue University\\
West Lafayette, IN, USA\\
            {\tt\small amanerik@purdue.edu}
\hspace{2cm}{\tt\small tprakash@purdue.edu}
\hspace{2cm}{\tt\small kak@purdue.edu}
}

\maketitle
\ifwacvfinal\thispagestyle{empty}\fi

\begin{abstract}

This work addresses the question whether it is possible to
design a computer-vision based automatic threat recognition
(ATR) system so that it can adapt to changing specifications
of a threat {\em without having to create a new ATR each
  time}. The changes in threat specifications, which may be
warranted by intelligence reports and world events, are
typically regarding the physical characteristics of what
constitutes a threat: its material composition, its shape,
its method of concealment, etc.  Here we present our design
of an AATR system (Adaptive ATR) that can adapt to changing
specifications in materials characterization (meaning
density, as measured by its x-ray attenuation coefficient),
its mass, and its thickness.  Our design uses a two-stage
cascaded approach, in which the first stage is characterized
by a high recall rate over the entire range of possibilities
for the threat parameters that are allowed to change.  The
purpose of the second stage is to then fine-tune the
performance of the overall system for the current threat
specifications. The computational effort for this
fine-tuning for achieving a desired PD/PFA rate is far less
than what it would take to create a new classifier with the
same overall performance for the new set of threat
specifications.
\end{abstract}


\section{Introduction}

Automatic threat recognition (ATR) systems for applications
such as airport passenger baggage screening are subject to
expensive and time-consuming processes of certification
before they can be deployed.  As is to be expected, such
systems are designed for a particular set of threat
specifications.  Unfortunately, the real world being what it
is, the precise specifications of a threat do not remain
constant with time and depend much on the world events and
the evolving capabilities of the bad guys out there.

Since the cost of developing a totally new ATR for a new set
of specifications can be expensive and time consuming, it
is necessary to explore the possibilities related to the
design of adaptive automatic threat recognition (AATR)
systems that can be quickly adapted to changing threat
specifications. \footnote{The need and technical requirements for AATR were developed by Carl Crawford (Csuptwo), Harry Martz (Lawrence Livermore National Laboratory) and Laura Parker (DHS Explosives Division Science and Technology Directorate) in collaboration with Northeastern University's Awareness and Localization of Explosives-Related Threats (ALERT) Center, a DHS Center of Excellence.  The datasets and scoring tools used in the paper were provided by ALERT, which was funded by DHS.}

In the context of airport baggage screening 
using 3D imaging based on X-ray tomography, threats like
home-made explosives (HMEs) \cite{wells2012review} and
firearms are characterized by parameters such as materials
and their composition, their shapes, the methods expected to
be used for their concealment, and so on. When the
specifications of such threats change, the modifications are
generally with respect to these parameters. In this paper,
we will focus exclusively on making an ATR system for HME
detection adaptive with respect to changes in the materials
density (as measured by its x-ray attenuation coefficient),
its thickness, and its mass.

The adaptive framework we present in this paper employs a
two-stage classifier cascade in which the first stage is
designed to operate with a sufficiently high recall rate
over the entire range of expected variability in the threat
parameters that could change.  The second stage classifier
is then fine-tuned to the parameters of the current threat
specification.

The two-stage classifier cascade is also provided with an
{\em adaptation protocol} for each threat parameter with
respect to which classifier adaptivity is desired.  This
protocol takes the classifier created for a given
specification of the parameters and modifies it to suit
another specification.  Depending on the nature of the
parameter involved, this protocol may entail a revisit to
the training data.  Even when the training data is revisited
for adaptation, the amount of work involved in the
adaptation process is far less than what it would take to
create a brand new classifier for the new set of parameter
specifications.

Our AATR system was tested on a dataset \cite{alertdataset}
made available by the DHS sponsored ALERT center at
Northeastern University specifically for the purpose of
evaluating AATR algorithms for airport baggage screening. A
unique feature of this dataset are its ORS (Object
Requirements Specification) files.  These files express the
specifications of a threat in terms of its mass, density and
thickness.  We demonstrate the ``adaptive power'' of our
approach by 
adapting our classifier to range of ORS files.

The organization of the paper is as follows: Section II provides a brief review of the existing work related to ATR systems and the response of prevalent ATR systems to varying parameter specifications. This is followed by a formal description of an Adaptive ATR system in Section III as well as the principle of operation behind our proposed approach. Section IV then introduces the  Cascaded Classification Approach to AATR system design encompassing the two-stage classifier model, the adaptation protocols adopted for the threat parameters and the training methodology with the adaptation protocols using the technique of Dynamic Sample Weighting. The implemented AATR system is then described in detail in Section V explaining the overall system operation. Finally, the implementation results for testing adaptability on different threat parameters are tabulated and illustrated in Section VI.

\section{Related Work}

ATR based on CT imaging for airport baggage inspection is
made challenging by the artifacts that result from metallic
objects that can be in arbitrary locations in a bag
\cite{karimi2015metal} \cite{xue2009metal}; by a lack of 
apriori structural information as compared to medical 
applications of CT \cite{mouton2013experimental}; and by 
large variability in the CT density range among the objects 
found in bags.

Much work on algorithms for automated baggage inspection has
been carried out under the auspices of the Department of
Homeland Security's ALERT (Awareness and Localization of
Explosive-Related Threats) initiatives on ATR segmentation
and object classification \cite{alertdataset}
\cite{alertto4initiative}. This has led to the development 
of a number of ATR segmenters and classifiers that could 
potentially be incorporated in the  airport checkpoint 
security pipeline in the future. Amongst these, the 
Stratovan Tumbler proposed by Wiley et al. 
\cite{wiley2012automatic} makes use of a 3D flood-fill
region-growing technique for segmenting out target blobs
from the query image. The method proposed by Grady et
al. \cite{grady2012automatic}, on the other hand, employs
isoperimetric graph partitioning
\cite{grady2006isoperimetric} to perform segmentation for
ATR. Song \cite{song2015vendor} proposed a sequential
segmentation and carving method for ATR segmentation
employing splitting and merging techniques for target 
extraction. Other proposed techniques use sieve 
decomposition algorithms and 
adaptive region growing for ATR.

More recent ALERT initiatives \cite{alertto4initiative} have
focused on a contextual classification of the object blobs
leading to the inception of more complex ATR systems using
multiple stages of segmentation and classification. Several
algorithms and ATR structures have been proposed in this
direction that include graph-based segmenters, MRF-EM based
image segmentation and decision-tree based ATR systems
\cite{alertto4initiative}. Other methods also provide for an
extension of ATR systems to dual-energy CT scans
\cite{mouton2015materials} as well for joint metal artifact
reduction and segmentation \cite{jin2015joint}.

For all these cases, the performance of the ATR systems is
evaluated on their ability to achieve desired values of
precision and recall for specific threats and fixed values
of threat parameters --- little analysis is made on what the
response of these systems would be if these threats and
threat parameters were to vary during runtime. This would be
especially difficult for ATR methods described in
\cite{wiley2012automatic} and \cite{grady2012automatic}
wherein the segmentation routines are pre-tuned for each
threat specification. On the other hand, the ATR classifiers
in \cite{song2015vendor}\cite{alertto4initiative}
\cite{jin2015joint} can be re-trained to adapt to a new 
threat but this involves retraining the ATR from a scratch.

This paper thus analyzes the problem of building an Adaptive 
ATR system, i.e., the problem of desensitizing an ATR system 
to variations in threat specifications. The two-stage 
classifier and the adaptation protocols proposed in the 
paper present a modular structure to carry out this 
desensitization with respect to specific threat parameters 
(density, mass and thickness) and without resorting to a 
complete rebuilding of the ATR. The detailed  implementation 
of this model is elaborated in the following sections. 

\section{System Overview - Adaptive ATR}

\subsection{Problem Statement}
X-ray based threat recognition for airport baggage screening can be tricky in case of threats such as home-made explosives (HMEs) \cite{singh2003explosives} which do not conform to a distinct shape or form and can be easily concealed. Such threats are detected on the basis of a materials-based characterization that involves the Region-of-Responsibility (ROR) \cite{martz2009overview} or density range (in terms  of its x-ray attenuation co-efficient), total object mass and thickness amongst other parameters. These specifications are, however, subject to frequent updates and modification with newer developments and alerts and therefore also require the deployed ATR systems  to be adaptive to these updates. Adjusting and re-certifying the ATR system for newer updates, however, is time-consuming and may result in an undesirable operational downtime. This puts forth the notion of the design of an Adaptive ATR (AATR) system, which is characterized by the ability to handle modifications to threat specifications during runtime while taking minimal effort to retrain or reconstruct the ATR. 

\subsection{Principle of Operation:}
In this paper, we present a two-stage classifier cascade architecture for the design of an AATR system - the focus for this ATR system design is exclusively with respect to the set of threat parameters that include the material density range, total object mass and object thickness. The two stage classifier structure makes use of the knowledge that while the precise specifications of these threat parameters may require modification during ATR operation, the range within which the parameters may vary can be pre-determined and remains fixed. Thus, the first stage of the cascade is designed to operate over this entire range with a high recall rate while the second stage fine-tunes the overall system performance by narrowing down the detection to the precise specification. Any modification to threat specifications therefore requires adjusting only the second stage of the cascade thus taking minimal effort to retrain the system. 

Now, ATR system performance is generally evaluated on the basis of the following metrics described in \cite{wells2012review}:
\begin{align}
& \textnormal{Probability of Detection, } PD 
&=& & \frac{TP}{TP+FN} \\
& \textnormal{Probability of False Alarm, } PFA 
&=& & \frac{FP}{FP+TN}
\end{align}

where TP, FP, TN and FN are True Positive, False Positive, True Negative and False Negative values respectively for the detected and total number of target objects on the dataset used for training the ATR. Here, Probability of Detection (PD) denotes the probability that the detected object is a threat while the Probability of False Alarm (PFA) denotes the probability that the detected object is a false alarm. We use these same performance metrics to evaluate the AATR system response as well - the proposed AATR design aims at obtaining PD and PFA values that are as close to the target values as possible.    

For each of the three threat parameters, namely, Density, Mass and Thickness, we employ an adaptation protocol specified by the Dynamic Sample Weighting technique explained in later sections. This technique allows tuning the system performance for the precise specification values of the threat parameters and also allows for adjusting system PD and PFA by moving along the ROC curve.

The dataset used for training and testing our architecture is the TO-4 dataset \cite{alertdataset} made available by ALERT center at Northeastern University - this dataset contains 188 instances of CT baggage scans with labeled ground truth images showing saline, rubber and clay objects. We used threat specifications expressed in terms of target ranges of density, mass and thickness to test the adaptive power of our system  - system performance was evaluated by creating an AATR  classifier for one specification set, creating new specification sets by modifying one or more threat parameter values and checking how well the new system adapts to these new specifications in terms of PD and PFA.  

\section{Proposed Methodology - Cascaded Classifier Architecture:} 
\begin{figure*}[th]
\centering
\includegraphics[scale=0.53]{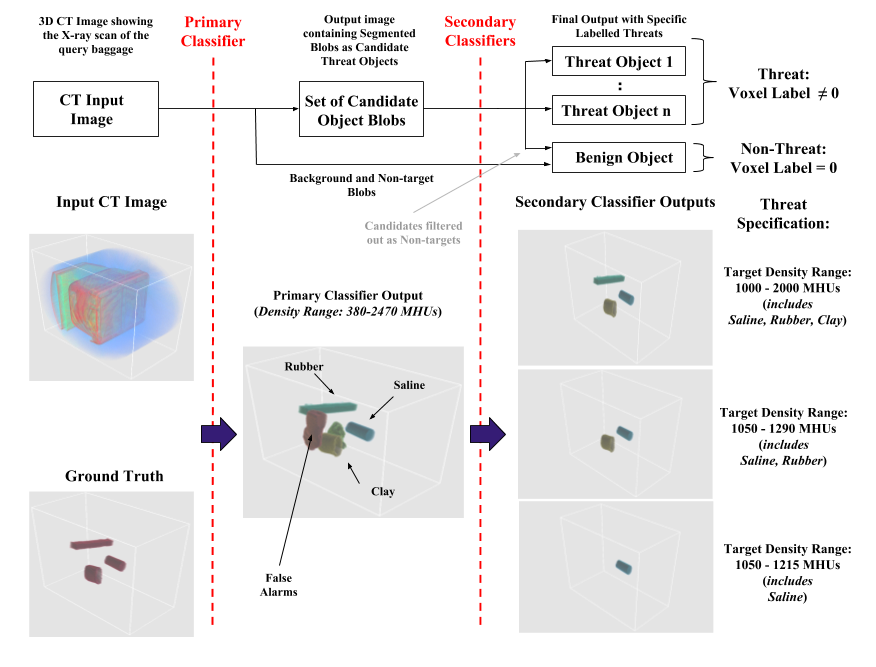}
\caption{\textit{Block Diagram for the Two-Stage Classifier Cascade Architecture for the AATR System with an Example Illustration}. The example shows the AATR operation for the case of three threat specifications with varying density ranges in Modified Hounsfield Units or MHUs. ($1000-2000$ MHUs, $1050-1290$ MHUs, $1050-1215$ MHUs). The Stage I classifier generates a set of candidate blobs over a wider density range ($380-2470$ MHUs) while the Stage II classifier then narrows down the detection to the specific density range.}
\label{bd_aatr}
\end{figure*}

Figure \ref{bd_aatr} shows the block diagram for the two-stage cascaded classifier AATR architecture along with an example illustration for the case of a varying density range specification for threat detection. (To characterize X-ray based detection, we express the density range in Modified Hounsfield Units or MHUs, i.e., Hounsfield Units offset by 1000 to yield a non-negative range of values). From the example illustration, we can see how the Stage I classifier which is designed to detect threats or targets over a large density range (380 - 2470 MHUs,  encompassing all possible densities of the target objects) generates an output with a high recall rate. The second classifier in the cascade then fine-tunes this result by narrowing down on the precise density range for the particular target specification (for example, 1050-1215 MHUs for saline objects). Changes in target specifications therefore only warrant a change in the second classifier which avoids training the entire AATR system for the new modification (e.g., changing to a new density range of 1170-1290 MHUs for detecting rubber objects). A stage-wise description of the system is given ahead:

\subsection{Stage-wise System Description:}

\subsubsection*{Stage I Classifier:}
The basis for operation of the cascaded classifier architecture lies in the notion that while the exact target specifications for ATR operation are subject to change, the overall range within which these values vary remains fixed. The Stage I classifier therefore is designed to identify any object whose threat specification falls within the acceptable global range for any of the threat parameters. Because this range remains fixed, the Stage I classifier does not need to be retrained or retuned for every new modification to the threat parameter values during the AATR operation. For the current implementation, the Stage I classifier is a Graph-based segmenter \cite{grady2012automatic} that segments out and generates a set of candidate target blobs from the input query image. Since all of these blobs do not satisfy the precise threat specification, the output of the Stage I classifier is riddled with a number of false alarms giving a high PFA rate - this is taken care of in the second stage of the cascade.

It is important to remember that the objective of an ATR system is not to partition the query image into threat and non-threat regions but to extract uniquely labeled segments from the image that correspond to different objects within the baggage. Thus, a challenging task for an ATR system in general is to be able to distinguish between two touching objects or one object contained within another - it is a non-trivial problem as one explosive concealed within a collection of similar but benign objects can be overlooked as a false alarm if not treated as a separate segment during the ATR operation. This problem is addressed in our implementation through the use of graph partitioning \cite{shi2000normalized}.

\subsubsection*{Stage II Classifier:}
Once the set of candidate target blobs has been generated by the Stage I classifier, the Stage II classifier in the cascade fine-tunes the result by classifying these blobs as per the exact threat specifications. For the case in Figure \ref{bd_aatr}, the Stage II classifier in the cascade filters only those blobs from the Stage I classifier output that match the density range provided in the threat specification. Depending on the threat parameter to be modified, the classifier may need to be retrained by revisiting the training data but since this constitutes only a part of the entire AATR cascade, it avoids retraining the entire system.  We employ different adaptation protocols for the threat parameters under consideration, each of which is explained in the next section.

\subsection{Adaptation Protocols:}
Diffferent threat parameters warrant different adaptation protocols to be adopted to adjust to a new modification. For parameters such as mass, this may be as simple as changing a threshold to prune lighter masses from the candidate blobs while for parameters like density, it may require retraining the entire Stage II classifier using histogram-based descriptors. The three different adaptation protocols for the three parameters under consideration, i.e., Mass, Density and Thickness are described as follows:

\subsubsection*{Tuning for Target Mass Range:}
As mentioned above, the protocol for adapting to a modification in the target mass range simply involves changing the threshold for pruning light masses from the set of candidate blobs generated by the Stage I classifier. A very good approximation of the mass of the candidate blob can be obtained by integrating over the voxel-wise CT density values within the blob volume - this can be used to filter out those blobs which do not fit into the specified mass range. 

\subsubsection*{Tuning for Target Density Range:}
Density range specifications for threat identification are the main parameters for the materials based characterization of the threat and are determined from the average density range of the constituent material of the target \cite{alertdataset}. However, since the material composition of any object is never completely homogeneous, it is difficult to determine the material composition by considering an absolute range for the density  specification \cite{wiley2012automatic} - this is especially difficult for small and thin blobs wherein the density distribution is corrupted by even a small quantity of noise, artifacts or contamination.

To adapt the AATR system to varying densities, therefore, we make use of a random forest classifier that is trained over the ALERT TO-4 dataset to identify target blobs composed of the desired material of interest. This classifier makes use of Normalized Density Histograms as classification features and is trained and tested on the dataset using ten-fold cross-validation. Specification of a new density range for the threat, thus, requires re-training of this classifier for the new desired density window. 

\subsubsection*{Tuning for Target Thickness Range:}
Thickness has become an important parameter for threat recognition since the occurrence of several recorded incidents of transporting HMEs concealed as thin plasticized sheets. Determining thickness of a sheet object can be difficult especially as the sheet within the baggage can be placed in a mangled or folded form. The adaptation protocol for the target thickness range involves construction of a 3D Thickness Vector for the candidate blob. This thickness vector is calculated for an object by calculating the median thickness of the object along each of its oriented principal axes and normalizing it over the largest thickness value (for the multiple folds of the sheet, the thickness vector is scaled by a suitable multiplier). A simple KNN Classifier based on this vector is then trained to identify objects within the target thickness range. In our implementation, the Thickness Vector is used for Dynamic Sample Weighting by concatenating it to the Density Histogram feature vector.

\subsection{Dynamic Sample Weighting:}

In our AATR implementation, the three adaptation protocols are implemented simultaneously using a single classifier through use of Dynamic Sample Weighting. This technique regulates the retraining of a classifier to adapt to the modification of one threat parameter by utilizing the knowledge of the values of all available parameters. By assigning a higher weight to those training samples whose parameter values are closer to the desired range and training the classifier accordingly, a better classification response can be obtained compared to the independent execution of the adaptation protocols.

In this method, a random forest classifier is constructed wherein the feature vector contains the Normalized Density Histogram and the Thickness Vector concatenated together.
While training, each training sample in the dataset is assigned a weight that is determined by the vicinity of the sample parameter values to the target parameter range. To calculate the weights, we take into account the specified range for each of the three threat parameters and construct a Gaussian Weighing Function (See Figure \ref{gww}). This can be exemplified for the case of Thickness as follows:
\begin{figure}[h]
\centering
\includegraphics[width=\linewidth]{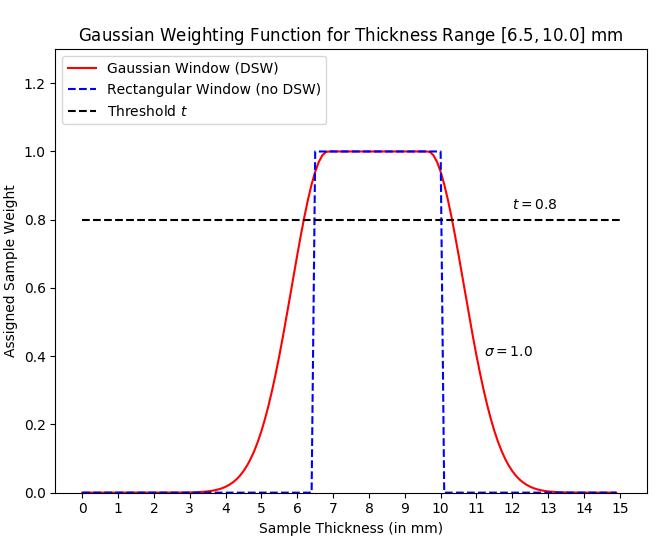}
\caption{ \textit{Gaussian Weighting Function} for Target Thickness Range $[6.5, 10.0]$ mm with a standard deviation, $\sigma=1.0$ and binarizing threshold, $t=0.8$. The parameters $\sigma, t$ are optimized to obtain the desired PD-PFA performance while retraining.}
\label{gww}
\end{figure}

Let us consider the thickness range specified by the minimum and maximum values of 6.5 mm and 10.0 mm respectively. Comparing the thickness of the current training sample with these limiting values, a sample weight is assigned to this sample using the Gaussian Weighting function illustrated in Figure \ref{gww}. The expression for this Gaussian Weighting Window for minimum and maximum limiting values, $T_{l}$ and $T_{h}$ is given by:
\begin{align}
w(t) &=& e^{\left[ \left( \frac{t - T_{l}-0.1T_d}{\sigma} \right)^2 \right]} & & t \leq  T_{l}'   \\  
     &=& 1.0   														& & T_{l}' < t  \leq  T_{h}' \nonumber\\ 
     &=& e^{\left[ \left( \frac{t - T_{h}+0.1T_d}{\sigma} \right)^2 \right]} & & T_{h}' \leq t \nonumber
\end{align}

where $\sigma$ is the standard deviation, t is the thickness of the current sample and $T_d=|T_{h}-T_{l}|$ is the thickness range, $T_h' =  T_{h}-0.1T_d$ and $T_l' =  T_{l}+0.1T_d $.

The total sample weight is thus the product of the individual sample weights for the threat parameters, Mass, Thickness and Density. The classifier is retrained with these sample weights for any modification in the specified threat parameters. Dynamic Sample Weighting also allows tuning the classifier performance  for a target PD and PFA - this is done by adjusting the standard deviation $\sigma$ of the Gaussian weighting function and the threshold $t$ on the sample weights that generates the positive/negative samples for the specific OOI. A grid-based search is used to find the optimum values of $\sigma, t$  that give the PD and PFA values closest to the target values during cross-validation.

\subsection{Tuning for Desired PD and PFA Response:}
\begin{figure}[h]
\centering
\includegraphics[width=0.95\linewidth]{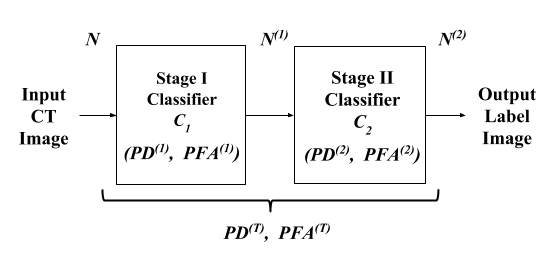}
\caption{\textit{Adjusting PD/PFA for the AATR system:} The two-stage cascade allows for a mechanism to adjust the target PD/PFA, $PD^{(T)}, PFA^{(T)}$ for the entire system by adjusting only the PD/PFA for $C_2$, i.e., $PD^{(2)}, PFA^{(2)}$. In the figure, $N^{(i)}$ denotes the set of filtered voxels at the $i^{th}$ stage;  $PD^{(i)}, PFA^{(i)}$ denote the PD/PFA for the $i^{th}$ classifier.}
\label{pdpfa_tune}
\end{figure}
To adapt to the desired PD/PFA requirements, the cascaded classification structure requires a mechanism which allows tuning the AATR classifier performance along the ROC curve without retraining the entire AATR system. For the cascaded architecture, this can be done by tuning the respective Stage II classifier as elaborated below:
\smallskip

Consider the Stage I classifier $C_1$ and the Stage II classifier $C_2$ connected in a cascade as shown in Figure \ref{pdpfa_tune} with the respective PD, PFA values $PD^{(1)}, PFA^{(1)}$ and $PD^{(2)}, PFA^{(2)}$  while the target PD, PFA for the total system are given by $PD^{(T)}, PFA^{(T)}$.
As the output of $C_1$ is directly fed to $C_2$ as input, it is evident that the total PD of the cascaded system is equivalent to the product of the PD's of the individual blocks and this holds true for PFA as well:
\begin{eqnarray}
PD^{(T)}  &=& PD^{(1)}\cdot PD^{(2)} \\
PFA^{(T)} &=& PFA^{(1)}\cdot PFA^{(2)} 
\end{eqnarray}

Now, because the Stage I classifier is only trained once, the values for $PD^{(1)}, PFA^{(1)}$ remain fixed but $PD^{(2)}, PFA^{(2)}$ can be adjusted for every new modification to the threat specifications as the Stage II classifier is retrained.
The target PD, PFA can thus be obtained by tuning $C_2$ to attain the following PD, PFA values:
\begin{equation}
PD^{(2)}  = \frac{PD^{(T)}} {PD^{(1)}}; \>\>\>\>\>\>\>
PFA^{(2)} = \frac{PFA^{(T)}} {PFA^{(1)}} 
\end{equation}

By using Dynamic Sample Weighting in our implementation, we attempt to attain the desired PD, PFA values by adjusting the thresholds of the sample weights and the spread of the Gaussian weighting function.

\section{System Implementation}
We have implemented the proposed two-stage cascaded classification structure using a Graph-based ATR segmenter and a Random Forest  classifier as the Stage I and Stage II classifiers in the cascade. The AATR system was tested for its adaptability by subjecting it to different sets of threat specifications grouped into three categories: \textit{ Varying Density (or constituent materials),  Varying Mass} and \textit{Varying Thickness} - the ALERT TO-4 dataset \cite{alertdataset} was used for the testing that included different target objects selected out of three Materials-of-Interest (MOIs), namely, \textit{saline, rubber} and \textit{clay}. The dataset contains 188 training samples consisting of CT scans of baggage that contained both target and non-target objects, in bulk as well as sheet form. The details for each stage of the structure are given ahead:

\subsection{Stage I Classifier:}

\begin{figure*}[h]
\centering
\includegraphics[width=11cm]{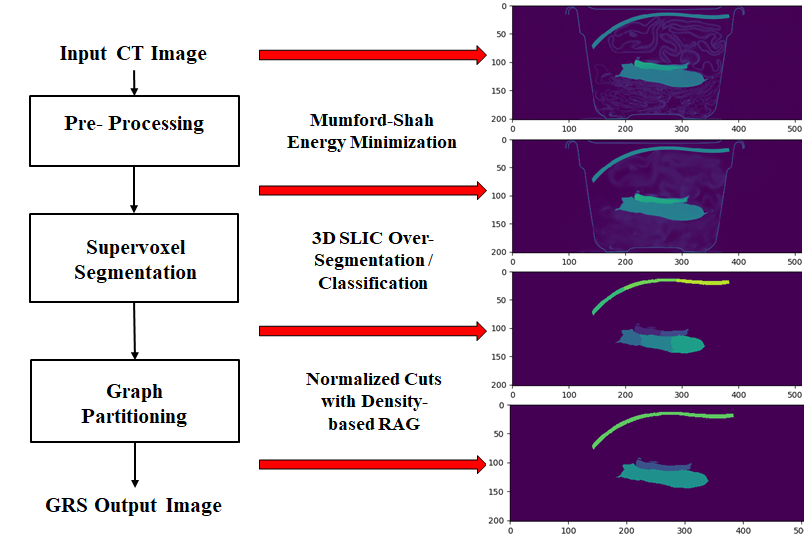}
\caption{\footnotesize Block Diagram for the Graph-Based Segmenter implemented as the Stage I Classifier. The segmenter works in three stages: (\textit{i) Pre-processing} - Mumford-Shah Energy Functional is used to reduce noise/artifacts, (ii) \textit{Supervoxel Segmentation} - Image is oversegmented to generate candidate supervoxels, (iii) \textit{Graph Partitioning} - Filtered supervoxels are partitioned on a density-based RAG (Region Adjacency Graph) to generate the candidate blobs.}
\label{grs_3}
\end{figure*}

The Stage I Classifier is constructed based on the ATR described in \cite{wiley2012automatic} with its  blocks explained below:

\subsubsection*{Pre-Processing:} The CT query image is first pre-processed using Mumford-Shah energy minimization \cite{mumford1989optimal} via the Ambrosio-Tortorelli approximation \cite{ambrosio1990approximation} to reduce the noise and artifacts and generate smooth sections for further segmentation. The 3D energy minimization is carried out with a parameter set for the minimizer $(\alpha, \beta, \epsilon)$ \cite{ambrosio1990approximation} adjusted to $(1000,0.9,0.1)$.

\subsubsection*{Supervoxel Segmentation:} The pre-processed image is then subjected to Supervoxel segmentation using SLIC \cite{achanta2012slic} with an initial number of segments $n\_segments=1000$ and compactness, $c=40$. The supervoxels are then filtered depending on whether their mean density lies within the pre-determined global range for threat densities ($380-2470$ MHUs). 

\subsubsection*{Graph-Partitioning:} 
To address the problem of segmenting out individual objects uniquely, a Normalized Cuts algorithm \cite{shi2000normalized} is used to partition the set of filtered supervoxels into a set of candidate blobs on the basis of intensity difference and the presence of a distinct boundary between two supervoxels. To do so, a graph is constructed for the set of supervoxels with edge weights calculated as follows:
\begin{equation}
w(i,j) = \eta_{e}\cdot\eta_{n}\cdot \exp \left[ \frac{(I_i-I_j)^2}{\sigma}\right]\exp \left[ \frac{S(i,j)^2}{0.25}\right]
\end{equation}
where:
\begin{itemize}
\item$\footnotesize
S(i,j) =\frac{\textnormal{  No. of boundary voxels with a distinct edge}  }{\textnormal{ Total No of boundary voxels between Nodes i, j}}$
\item $\eta_{n}$ - Set to 1 if Node i and Node j are neighbors
\item $\eta_{e}$ - Set to 1 if  $S(i,j) > 0.25$.
\item $I_i,I_j$ - Mean intensities of Nodes i and j.
\item $\sigma$ - Global density range for all threats.
\end{itemize}
The Normalized Cuts algorithm is then applied  with a threshold, $t=0.1$ to generate the set of candidate blobs.

\subsection{Stage II Classifier - Random Forest:}

For the threat parameters specified for a known constituent material, i.e., with samples present in the TO-4 dataset (namely, saline, rubber or clay), the Stage II classifier used is a Random Forest classifier with 50 estimators and using the Gini impurity criterion. 
The Stage II classifier uses a feature vector consisting of a 100-bin Normalized Density histogram and a 3D Thickness vector concatenated together. The density histogram features are weighted with a value of $\frac{0.5}{100}=0.005$ while the thickness vector is weighted with $\frac{0.5}{3}=0.167$ to normalize the feature vector - if the target thickness does not change in the new modification, the respective weights are set to zero. The output blobs of the Stage II classifier are then filtered using the mass feature pruning lighter masses from the final output image.  


\section{Results}
\begin{figure}

    \begin{subfigure}[b]{0.5\textwidth}
        \includegraphics[width=0.94\textwidth]{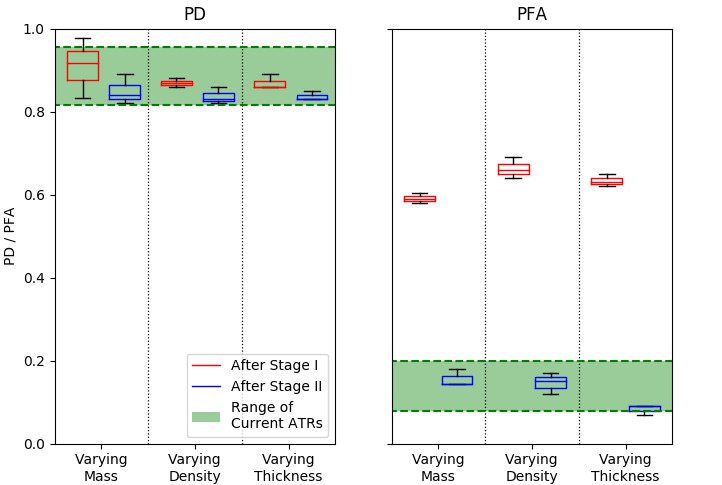}
        \caption{AATR 1*: Graph-based ATR Segmenter for Stage I}
    \end{subfigure}
    \begin{subfigure}{0.5\textwidth}
        \includegraphics[width=0.94\textwidth]{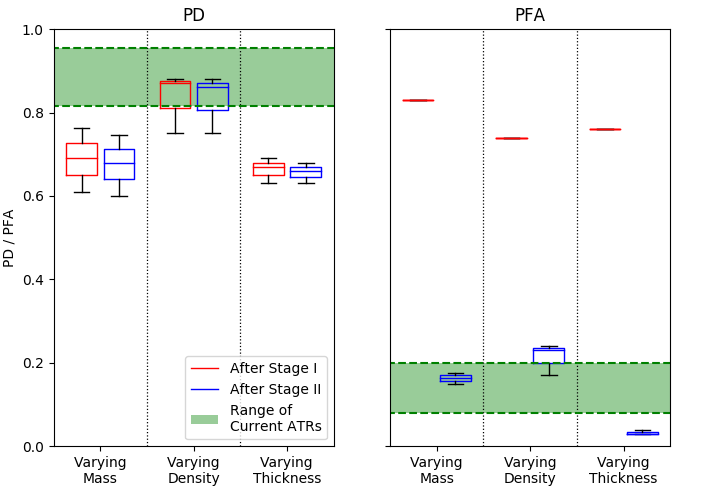}
        \caption{AATR 2: Supervoxel-based ATR for Stage I}
    \end{subfigure}
    \begin{subfigure}{0.5\textwidth}
        \includegraphics[width=0.94\textwidth]{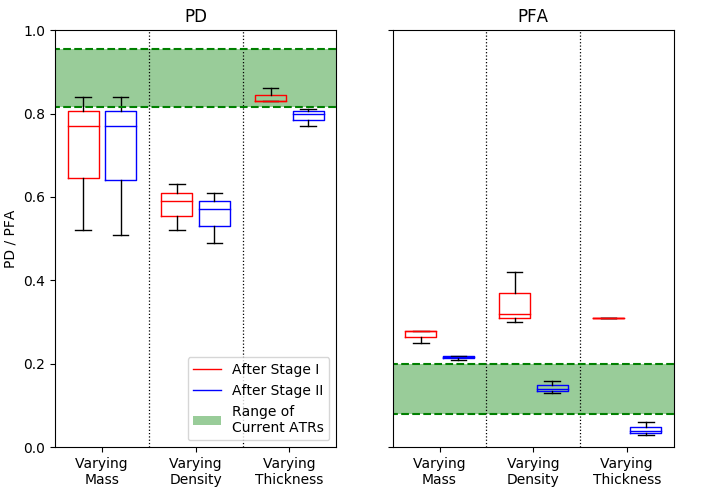}
        \caption{AATR 3: CCL-based Segmenter for Stage I}
    \end{subfigure}
    \caption{\textit{PD/PFA Performance of AATRs with different systems for Stage I classification:} 
    The figure shows the response of three different AATRs for the cases of Varying Density, Varying Mass and Varying Thickness. Each AATR is implemented with a different system as its Stage I classifier - these systems include a Graph-based ATR Segmenter (AATR 1), a Supervoxel-based ATR classifier (AATR 2) and a CCL-based ATR segmenter (AATR 3) respectively. The performance range of current ATRs for PD and PFA has been obtained from \cite{alertto4initiative} is denoted by a green band in the figure. *AATR 1 gives the best adaptive response and has been explained in detail in this paper. 
    }
\end{figure}

\begin{table}[h]
\centering
\caption{Adaptation to Varying Threat Parameters (AATR 1) - Varying Density Range (Target Materials)}
\label{tb:am2}
\begin{tabular}{ccc}
\hline
Material           & Stage I Classifier                & AATR   \\
(Density Range)    & (PD $\%$, PFA $\%$)   & (PD $\%$, PFA $\%$)  \\
\hline
 & & \\
saline$^*$  & (91 $\%$, 53 $\%$)           &  (90 $\%$, 12 $\%$)         \\
rubber$^*$  & (87 $\%$, 62 $\%$)           &  (85 $\%$, 13 $\%$)         \\
clay$^*$    & (83 $\%$, 66 $\%$)           &  (84 $\%$, 13 $\%$)         \\
\hline
\end{tabular}
\bigskip
\\
\caption{Adaptation to Varying Threat Parameters (AATR 1) - Varying Mass}
\label{tb:am4}
\begin{tabular}{ccc}
\hline
Mass         & Stage I Classifier             & AATR   \\
Range        & (PD $\%$, PFA $\%$) & (PD $\%$, PFA $\%$) \\
\hline
 & & \\
 $ >400$ g   & (86 $\%$, 69 $\%$)           & (83 $\%$, 12 $\%$)        \\
 $ >300$ g   & (88 $\%$, 66 $\%$)   	    & (86 $\%$, 17 $\%$)        \\
 $ >100$ g   & (86 $\%$, 64 $\%$)			& (82 $\%$, 15 $\%$)        \\
\hline
\end{tabular}
\bigskip
\\
\caption{Adaptation to Varying Threat Parameters (AATR 1) - Varying Thickness}
\label{tb:am5}
\begin{tabular}{ccc}
\hline
Thickness       & Stage I Classifier              & AATR             \\
 Range          & (PD $\%$, PFA $\%$)   & (PD $\%$, PFA $\%$) \\
\hline
 & & \\
$> 10.0\>\>\>$  mm   & (89 $\%$, 65 $\%$) 	        &   (85 $\%$, 7 $\%$)         \\
$6.5-10.0$ mm   & (86 $\%$, 63 $\%$)  	    &   (83 $\%$, 9 $\%$)         \\
$0-6.5\>\>\>\>$    mm   & (86 $\%$, 62 $\%$) 		    &   (83 $\%$, 9 $\%$)         \\
\hline
\end{tabular}
\\ 
\begin{justify}
\footnotesize \textbf{Note:} The \textit{saline, clay} and \textit{rubber} objects correspond to the density ranges (1050-1215), (1170-1290) and  (1530-1715) MHUs respectively. The difference between $PFA^{(T)}$ and $PFA^{(1)}$ for all cases shows the effect of fine-tuning the Stage I classifier output in the second stage of the cascade.  
\end{justify}
\end{table}

\begin{figure}[thbp]
\centering
  \includegraphics[width=0.45\textwidth]{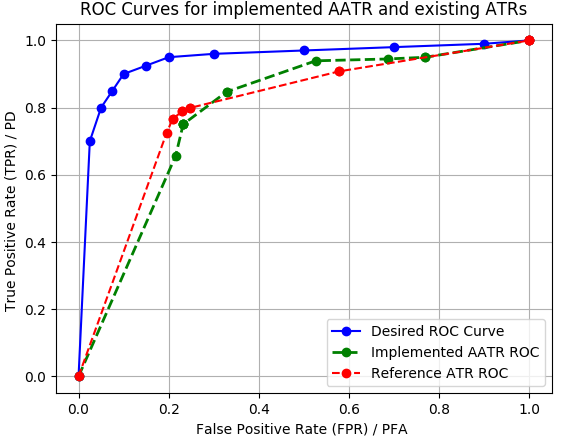}
\caption{\textit{ROC Response Curve for the Implemented AATR}: 
(i) the implemented AATR system, 
(ii) a Supervoxel-based ATR classifier for reference \cite{alertto4initiative}, 
(iii) the desired ROC curve for system operation. 
The systems were trained to detect saline objects with a fixed set of specifications to achieve the desired ROC curve. The ATR in this case was completely retrained for a new variation while the AATR adapts to the variations.}
\label{fig:results}
\end{figure}

The implemented AATR system was tested for its adaptability on the ALERT TO-4 dataset \cite{alertdataset}. This dataset consists of 188 CT scans of bags that contain target objects made from \textit{saline, rubber} and \textit{clay} and non-target objects that can typically be found in passenger baggage at airports.  The target objects are both in bulk and in sheet forms.

As previously mentioned in the Introduction, target specifications for testing for adaptability are in the form of ORS (Object Requirements Specification) files. Through a specification of the minimum and maximum value for the three parameters --- mass, density, and range --- an ORS file declares what it means for an object to be a target.  Each ORS file also specifies the {\em desired} PD and PFA for the classifier performance for that target.


We tested our AATR system for adaptability by subjecting it to nine ORS files, each with a different threat specification for one of the three material specific parameters --- mass, thickness, or density.  In three of the ORS files, only the mass specification changed, in another grouping of three, only the density changed, and in yet another grouping of three, only the thickness changed.  With regard to the desired AATR performance, it was the same for all ORS files: $90 \%$ for PD and $10 \%$ for PFA.

The PD/PFA performance as achieved by our AATR systems for the three ORS groupings is illustrated in Figure 5. The figure shows the PD/PFA values achieved for both the Stage I classifier (red) and the total AATR system (blue) after the Stage II classifier is added - this is shown for the case of three AATRs implemented with a different Stage I classifier: AATR 1 is implemented from a Connected Component Labeling (CCL)-based ATR segmenter \cite{alertto4initiative}, AATR 2 is derived from a Supervoxel-based ATR classifier while AATR 3 is based on the Graph-based ATR segmenter \cite{grady2012automatic} that we have implemented and explained in detail in the previous sections. We see a marked drop in the PFA values as a result of the fine-tuning performed by the Stage II classifier for all three cases.  For comparison, the green band in the figure is for the mainstream ATRs described in \cite{alertto4initiative}.  This figure establishes that the overall performance of an AATR framework need be no worse than what it is for well-designed ATR algorithms out there. This is also exemplified by a comparison between the ROC curves of the AATR system with that of an ATR system trained on a fixed set of material-specific parameters, as shown in Figure 6  - our use of Dynamic Sample Weighting produces a response within an acceptable range from the ATR ROC curve.


The PD/PFA response for AATR 1 is also tabulated in Tables \ref{tb:am2},
\ref{tb:am4} and \ref{tb:am5} for a numerical
demonstration. We see that the standard deviations for the
PD and PFA do not exceed above $2.62 \%$ and $2.05 \%$
respectively in all cases showing stable behavior against
varying threat specifications.

\section{Conclusion}

The results obtained with our AATR architecture for
designing adaptive classifiers for threat recognition show
that the overall classification performance with an adaptive
framework need be no worse than what can be achieved with a
traditional approach that calls for creating a brand new
classifier for each new definition of what constitutes a
threat.  Considering that the work required for the
fine-tuning of the Stage II classifier in our AATR is much
less than what it would take to create an ATR from ground
zero, our work establishes the viability of AATR frameworks
for automatic target recognition.


\begin{thebibliography}{10}\itemsep=-3pt
\small
\bibitem{alertdataset}
{ALERT (Awareness and Localization of Explosive-Related Threats) ATR Dataset}.
\newblock
  \url{http://www.northeastern.edu/alert/transitioning-technology/alert-datasets/},
  2014.

\bibitem{alertto4initiative}
{ALERT TO-4 ATR Initiative}.
\newblock
  \url{https://myfiles.neu.edu/groups/ALERT/strategic_studies/TO4_FinalReport.pdf},
  2014.

\bibitem{achanta2012slic}
R.~Achanta, A.~Shaji, K.~Smith, A.~Lucchi, P.~Fua, and S.~S{\"u}sstrunk.
\newblock Slic superpixels compared to state-of-the-art superpixel methods.
\newblock {\em IEEE transactions on pattern analysis and machine intelligence},
  34(11):2274--2282, 2012.

\bibitem{ambrosio1990approximation}
L.~Ambrosio and V.~M. Tortorelli.
\newblock Approximation of functional depending on jumps by elliptic functional
  via t-convergence.
\newblock {\em Communications on Pure and Applied Mathematics},
  43(8):999--1036, 1990.

\bibitem{grady2006isoperimetric}
L.~Grady and E.~L. Schwartz.
\newblock Isoperimetric graph partitioning for image segmentation.
\newblock {\em IEEE transactions on pattern analysis and machine intelligence},
  28(3):469--475, 2006.

\bibitem{grady2012automatic}
L.~Grady, V.~Singh, T.~Kohlberger, C.~Alvino, and C.~Bahlmann.
\newblock Automatic segmentation of unknown objects, with application to
  baggage security.
\newblock In {\em Computer Vision--ECCV 2012}, pages 430--444. Springer, 2012.

\bibitem{jin2015joint}
P.~Jin, D.~H. Ye, and C.~A. Bouman.
\newblock Joint metal artifact reduction and segmentation of ct images using
  dictionary-based image prior and continuous-relaxed potts model.
\newblock In {\em Image Processing (ICIP), 2015 IEEE International Conference
  on}, pages 798--802. IEEE, 2015.

\bibitem{karimi2015metal}
S.~Karimi, H.~Martz, and P.~Cosman.
\newblock Metal artifact reduction for ct-based luggage screening.
\newblock {\em Journal of X-ray science and technology}, 23(4):435--451, 2015.

\bibitem{martz2009overview}
H.~Martz and C.~Crawford.
\newblock Overview of deployed eds technologies.
\newblock Technical report, Lawrence Livermore National Lab.(LLNL), Livermore,
  CA (United States), 2009.

\bibitem{mouton2015materials}
A.~Mouton and T.~P. Breckon.
\newblock Materials-based 3d segmentation of unknown objects from dual-energy
  computed tomography imagery in baggage security screening.
\newblock {\em Pattern Recognition}, 48(6):1961--1978, 2015.

\bibitem{mouton2013experimental}
A.~Mouton, N.~Megherbi, K.~Van~Slambrouck, J.~Nuyts, and T.~P. Breckon.
\newblock An experimental survey of metal artefact reduction in computed
  tomography.
\newblock {\em Journal of X-ray Science and Technology}, 21(2):193--226, 2013.

\bibitem{mumford1989optimal}
D.~Mumford and J.~Shah.
\newblock Optimal approximations by piecewise smooth functions and associated
  variational problems.
\newblock {\em Communications on pure and applied mathematics}, 42(5):577--685,
  1989.

\bibitem{shi2000normalized}
J.~Shi and J.~Malik.
\newblock Normalized cuts and image segmentation.
\newblock {\em IEEE Transactions on pattern analysis and machine intelligence},
  22(8):888--905, 2000.

\bibitem{singh2003explosives}
S.~Singh and M.~Singh.
\newblock Explosives detection systems (eds) for aviation security.
\newblock {\em Signal processing}, 83(1):31--55, 2003.

\bibitem{song2015vendor}
S.~M. Song, B.~Kauke, and D.~P. Boyd.
\newblock Vendor and scanner independent common workstation for security,
  July~28 2015.
\newblock US Patent 9,094,580.

\bibitem{wells2012review}
K.~Wells and D.~Bradley.
\newblock A review of x-ray explosives detection techniques for checked
  baggage.
\newblock {\em Applied Radiation and Isotopes}, 70(8):1729--1746, 2012.

\bibitem{wiley2012automatic}
D.~F. Wiley, D.~Ghosh, and C.~Woodhouse.
\newblock Automatic segmentation of ct scans of checked baggage.
\newblock In {\em Proceedings of the 2nd International Meeting on Image
  Formation in X-ray CT}, pages 310--313, 2012.

\bibitem{xue2009metal}
H.~Xue, L.~Zhang, Y.~Xiao, Z.~Chen, and Y.~Xing.
\newblock Metal artifact reduction in dual energy ct by sinogram segmentation
  based on active contour model and tv inpainting.
\newblock In {\em Nuclear Science Symposium Conference Record (NSS/MIC), 2009
  IEEE}, pages 904--908. IEEE, 2009.

\end{thebibliography}

\end{document}